\documentclass[twocolumn,10pt,a4paper]{article}

\usepackage{graphicx}
\usepackage{float} 
\usepackage{CJKutf8} 
\usepackage{tikz} 
\usepackage{amsmath}
\usepackage{multirow}
\usepackage{hyperref}
\usepackage{url}
\usepackage{authblk}
\usepackage[top=25mm,bottom=30mm,left=25mm,right=25mm]{geometry}
\usepackage{enumerate}
\usepackage{caption}
\usepackage{cite}
\usetikzlibrary{arrows.meta}
\usetikzlibrary{fit}

\makeatletter
\renewcommand{\@makefntext}[1]{%
    \noindent
    \makebox[0pt][r]{\@thefnmark\ }#1

    \addvspace{6pt}
}
\makeatother

\title{Computational Implementation of a Model of Category-Theroretic Metaphor Comprehension}

\author[1]{Fumitaka Iwaki}
\author[2]{Miho Fuyama}
\author[3]{Hayato Saigo}
\author[1]{Tatsuji Takahashi}
\affil[1]{School of Science and Technology, Tokyo Denki University}
\affil[2]{College of Letters, Ritsumeikan University}
\affil[3]{ZEN University}
\date{}

\begin{document}

\renewcommand{\thefootnote}{}  

\maketitle

\footnotetext[1]{%
    F. Iwaki \\
    Ishizaka, Hatoyama, Hiki, Saitama, 350-0394, JAPAN \\
    E-mail: 25udj01@ms.dendai.ac.jp
}
\footnotetext[2]{%
    M. Fuyama \\
    56-1, Toujiin Kitamachi Kita-ku, Kyoto-Shi, Kyoto, 603-8577, JAPAN \\
    E-mail: mifuyama@fc.ritsumei.ac.jp
}
\footnotetext[3]{%
    H. Saigo \\
    3-12-11, Shinjuku, Zushi, Kanagawa, 249-0007, JAPAN \\
    E-mail: hayato\_saigo@zen.ac.jp
}
\footnotetext[4]{%
    T. Takahashi \\
    E-mail: tatsuji.takahashi@gmail.com
}
\renewcommand{\thefootnote}{\arabic{footnote}}  
\setcounter{footnote}{0}

\begin{abstract}
In this study, we developed a computational implementation for a model of metaphor comprehension based on the theory of indeterminate natural transformation (TINT) proposed by Fuyama et al.
We simplified the algorithms implementing the model to be closer to the original theory and verified it through data fitting and simulations.
The outputs of the algorithms are evaluated with three measures: data-fitting with experimental data, the systematicity of the metaphor comprehension result, and the novelty of the comprehension (i.e. the correspondence of the associative structure of the source and target of the metaphor).
The improved algorithm 
outperformed the existing ones in all the three measures.

\textbf{Keywords:} 
analogy; category theory; natural transformation; functor; theory of indeterminate natural transformation (TINT)
\end{abstract}

\section{Introduction}
The creativity of the human brain is exemplified by the ubiquity and centrality of metaphor in our thoughts
\cite{Chater-2019-MindFlata}.
The recent, dramatic progress of artificial intelligence has not been, however, due to replicating the flexibility in human imagination.
The intelligence of computers has been acquired by absorbing a vast number of problems solved in the past.
On the other hand, the key of human intelligence would be to find patterns where there seems to be no structure.
Modern artificial intelligences has not yet achieved finding correspondences or metaphors in and between the complex physical and psychological domains.
Understanding metaphor creation and comprehension processes is thus important for revealing the keys for flexible cognition.

Category theory, which can mathematically formulate structureal correspondences, has been applied in various fields outside mathematics in recent years (see e.g. \cite{Fong-2019-InvitationAppliedCategoryTheorya}).
Some researchers expect that the tools of category theory such as functors and natural transformations, can contribute to elucidating the mechanisms of deft transfer learning and analogical reasoning of humans~\cite{Ikeda-2021-ComputationalImplementationMetaphorComprehension}.
The theory of indeterminate natural transformation (TINT) which is based on category-theoretic concepts was proposed as a hypothesis for exploring the process of metaphor comprehension and creating meanings \cite{Fuyama2020a}.
Ikeda et al. computationally implemented TINT as algorithms and verified how well it describes human metaphor interpretation data \cite{Ikeda-2021-ComputationalImplementationMetaphorComprehension}.

TINT takes an \textit{analogy position} for metaphor comprehension~\cite{Holyoak-2018-MetaphorComprehensionCriticalReviewa},
considering a comprehension of metaphor as building a mapping from the semantic structure of the source of a metaphor to that of the target.
A representative model and algorithm of this position is structure-mapping theory~\cite{Gentner-1983-StructuremappingTheoreticalFrameworkAnalogya} and structure-mapping engine~\cite{Falkenhainer-1989-StructuremappingEngineAlgorithmExamples}, respectively.
While they assume the well-organized knowledge structures (tree structures with labeled predicate relations) for the analogical reasoning (see \cite{Chalmers-1992-HighlevelPerceptionRepresentationAnalogya} for a criticism),
TINT merely assumes a network structure of association, which is even `weaker' than the vector space embeddings of words, hence is potentially more general.

In TINT, the meaning of a word (in general, an `image') is represented by a `coslice category' which is the local associative structure around the image.
The comprehension process of a metaphor such as `Butterflies are dancers' is described by the interaction between the coslice categories of the two images, `butterfly' (target) and `dancer' (source).
In the implementation by Ikeda et al., the comprehension-correspondences between images are searched deterministically (using the argmax operator), which is not realistic considering the nature of association and multistability of human output.
Also, their implementation had some problems in terms of the type of the output correspondences that made the interpretation of the results difficult and the comparison by quantitative evaluation impossible.
In this paper, we introduce new algorithms implementing TINT that solves these problems.
As for the evaluation of the algorithms,
we test the data-fit according to the data and methods introduced in Ikeda et al.
We evaluate the algorithms in terms of a kind of systematicity of the comprehension-correpondences and the novelty of them.

\section{Theory of Indeterminate Natural Transformation}

\subsection{The Concept of TINT}
TINT is the theory that describes the process of understanding metaphors 
as the interaction between semantic structures of the source 
and the target
~\cite{Fuyama2020a}.
In TINT, the meaning of the images is the whole relationship around the images.
The theory describes the interaction between these structures of images.

In this case, the creation of a new meaning is considered to be the creation of new associative relationships between images.
This corresponds to the process of understanding a new metaphor in which the source and the target, which were not related before, are related by the metaphorical relationship.
For example, when the metaphor `Butterflies are dancers' is given, the association from `butterfly' to `dancer' appears.
Then the meaning of `butterfly' is updated by this change.
In this way, understanding a new metaphor can cause creation of a new meaning.
TINT attempts to better formalize the interaction between the meanings as the local structures around the images.

\subsection{Basic Concepts of Category Theory}

Here we introduce four basic notions of category theory used in formalizing TINT: category, functor, natural transformation, and coslice category\footnote{For an illustrated exposition of the category-theoretic concepts and their importance in cognitive science, see \cite{Fuyama2020a}}.
A \textit{category} $\mathcal{C}$ consists of objects $\mathrm{obj}(\mathcal{C})$ and arrows $\mathrm{arr}(\mathcal{C})$. Each arrow $f\in \mathrm{arr}(\mathcal{C})$ has the domain $\mathrm{dom}(f)=X \in \mathrm{obj}(\mathcal{C})$ and the codomain $\mathrm{cod}(f)=Y\in \mathrm{obj}(\mathcal{C})$, and denoted as $f: X\rightarrow Y$.
Arrows $f, g$ can be composed to $g \circ f$ under some condition ($\mathrm{cod}(f)=\mathrm{dom}(g)$), and satisfy the associative and unit laws (the units for composition are identity arrows such as $1_X$ which exist uniqutely for each object $X$).

A \textit{functor} $F: \mathcal{C}\rightarrow \mathcal{D}$ is a structure-preserving mapping between category $\mathcal{C}$ and $\mathcal{D}$, which maps each object/arrow in $\mathcal{C}$ to the corresponding object/arrow in $\mathcal{D}$, satisfying three conditions: 
(1) It maps $f:X \longrightarrow Y$ in $\mathcal{C}$ to $F(f):F(X) \longrightarrow F(Y)$ in $\mathcal{D}$.
(2) $F(f \circ g) =F(f) \circ F(g)$ for any composable pair of $f,g$ in $\mathcal{C}$.
(3) For each $X$ in $\mathcal{C}$, $F(1_X) = 1_{F(X)}$.

Let $F,G$ be functors from category $\mathcal{C}$ to category $\mathcal{D}$. 
A \textit{natural transformation} $\vartheta$
from $F$ to $G$ satisfies the following conditions:
(1) $\vartheta$ maps each object $X$ in $\mathcal{C}$ to the corresponding morphism $\vartheta_X:F(X)  \longrightarrow G(X)$ in $\mathcal{D}$.
(2) For any $f:X \longrightarrow Y$ in $\mathcal{C}, \vartheta_Y \circ F(f) = G(f) \circ \vartheta_X.$

\textit{Coslice category} $X\backslash \mathcal{C}$, where $X$ is an object in $\mathcal{C}$, is defined as follows.
Objects in $X\backslash \mathcal{C}$ are morphisms $f, g, ...$ in $\mathcal{C}$ from $X$ ($\mathrm{dom}(f)=\mathrm{dom}(g)=\cdots = X$).
A morphism between two objects $f_1:X\longrightarrow X_1$ and $f_2:X\longrightarrow X_2$
is the triple of $(f_1,f_2,g)$, where 
$g:X_1\longrightarrow X_2$ and $f_2=g\circ f_1$. 
In other words, the objects and arrows of $X\backslash \mathcal{C}$ are arrows from $X$ and commutative triangle diagrams, respectively.

\subsection{Modeling TINT via Category Theory}

\begin{figure}[tb]
    \centering
    \includegraphics[width=.6\linewidth]{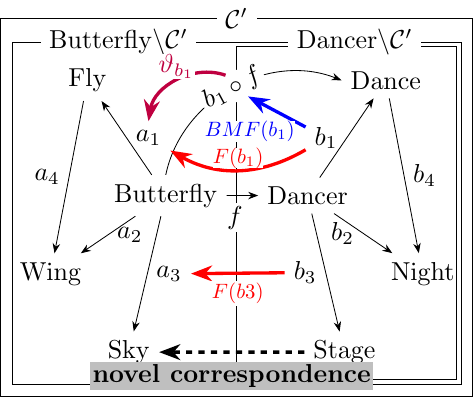}
    \caption{The specific example of an exploration in TINT, beginning with ``A butterfly is like a dancer.''}
    \label{fig:tint_specific_exp}
\end{figure}

TINT defines two categories, the \textit{latent} category of images, $\mathcal{C}$, and the \textit{elicited} category of associations
.
The latent category is a thin category (up to one arrow from an object to another) with an additional structure which is that each arrow $f$ has a weight $\mu_f \in [0,1]$.
As stated in Experiment, it can be a weighted complete directed graph on images.
The elicited category is also a normal thin category with no weight.
First, we define the category of images $\mathcal{C}$ as the semantic network where metaphor comprehension happens.
For an object $X$ in $\mathcal{C}$, the `meaning' of $X$ is represented by the coslice category $X\backslash \mathcal{C}$.
We model the process of metaphor comprehension as exploration of functors (structure-preserving mappings) between coslice categories (the meanings of the target and the source of a metaphor) using natural transformations, as follows.
Here we take as example of comprehension a metaphor `Butterflies are dancers' as in Figure \ref{fig:tint_specific_exp}

\begin{figure}[tb]
    \centering
    \includegraphics[width=.5\linewidth]{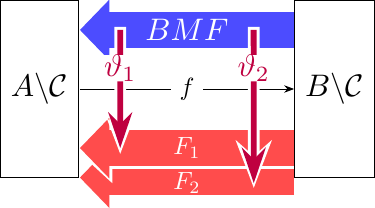}
    \caption{Schematic diagram of natural transformation search in TINT. Based on the canonical functor derived from the occurrence of association $f$, new functors such as $F_1$ and $F_2$ are searched as the construction of natural transformations $\vartheta_1$ and $\vartheta_2$, respectively.}
\end{figure}

\begin{enumerate}[a.]
  \setlength{\itemsep}{0pt}
  \setlength{\parskip}{0pt}
  \setlength{\parsep}{0pt}
  \setlength{\topsep}{0pt}
    \item
      The metaphor `Butterflies are dancers' is given, and the association from image `butterfly' to `dancer' occurs (an arrow $f: \mathrm{`butterfly'} \rightarrow \mathrm{`dancer'}$ is added to $\mathcal{C}$, if it did not exist).
      Category $\mathcal{C}$ may change by this arrow $f$, and we denote it by $\mathcal{C}'$.
    \item 
      Due to $f$, the unique functor $f\backslash \mathcal{C}'$ from $B\backslash \mathcal{C}'$ to $A\backslash \mathcal{C}'$ is created.
      It maps an arrow $b_i$ in $B\backslash \mathcal{C}'$ to $b_i \circ f$ in $A\backslash \mathcal{C}'$ (composition functor) and is called the base-of-metaphor functor (BMF).
      It is that, from associations
      `Dancer $\rightarrow$ Night'
      and
      `Dancer $\rightarrow$ Dance',
      BMF creates
      `Butterfly $\rightarrow$ Dancer $\rightarrow$ Night'
      and
      `Butterfly $\rightarrow$ Dancer $\rightarrow$ Dance'
      BMF is not clear in the interpretation nor non-trivial as 
      an interpretation of a metaphor.
    \item 
      Search for a new functor modeling a more natural metaphor interpretation by constructing a natural transformation from BMF.
      The two actual exploration procedures are described in the next section.
      This is equivalent to searching for a relationship (association) which
      corresponds to `Butterfly $\rightarrow$ Dancer $\rightarrow$ Dance',
      such as `Butterfly $\rightarrow$ Fly',
      and the correspondence from the former to the latter is a component of
      the natural transformation $\vartheta_{b_1}$.
      This leads to the discovery of a new functor $F$ which is defined by
      $\vartheta: \mathrm{BMF} \Rightarrow F$.
    \item 
      $F$ represents a correspondence such as `to dance for dancer is to fly for butterfly' and `stage for dancer is sky for butterfly'.
      As a result, images directly associated with $B$ and that of $A$ correspond to each other, and the meaning of the metaphor is interpreted.
\end{enumerate}

\subsection{TINT Algorithms}

\cite{Ikeda-2021-ComputationalImplementationMetaphorComprehension} proposed
  two algorithms that implement TINT.
One is an \textit{object-based} algorithm which considers only the objects of the coslice category and corresponds only the objects of each coslice category.
It does not consider the relationships (the commutative arrows of the coslice category).
The other is \textit{relation-based} in that it performs correspondence considering the commutative triangle structure.
The triangle structure is a triplet which is an arrow of the coslice categories.
These algorithms require the associative probabilities between images.
Ikeda et al. conducted cognitive experiments asking people about the associative strength between images and set the results as the weights on the arrows in the latent category $\mathcal{C}$.
In the following, the images associated from a certain image $X$ (that become the codomains of the arrows of the coslice category $X\backslash \mathcal{C}$) the initial images.

\subsubsection{The Object-based Exploration}

\begin{figure}[tb]
    \centering
    \includegraphics[width=\linewidth]{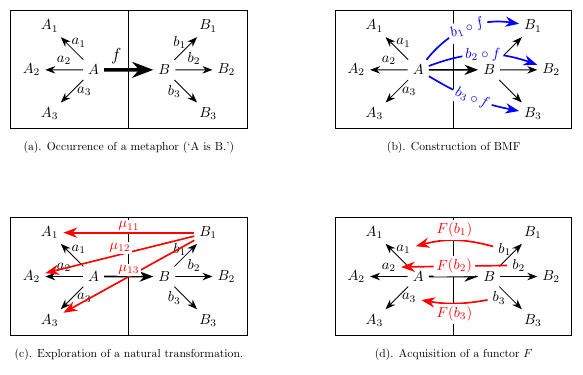}
    \caption{Construction of a natural transformation (at the same time a functor) by the object-based method}
    \label{fig:objective}
\end{figure}

\begin{figure}[tb]
    \centering
    \includegraphics[width=.6\linewidth]{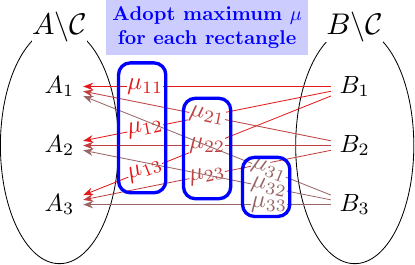}
    \caption{A correspondence between images via the objective method}
    \label{fig:tint_objective_selection}
\end{figure}

Here we explain the algorithm for exploring natural transformation/functor
based directly on the associations between images. 
The algorithm is described in four steps (from \ref{fig:objective}-a to \ref{fig:objective}-d) as in Figure \ref{fig:objective}.

\begin{enumerate}[\ref{fig:objective}-a]
\setlength{\itemsep}{0pt}
\setlength{\parskip}{0pt}
\setlength{\parsep}{0pt}
\setlength{\topsep}{0pt}
\item 
    Set the initial state of the coslice categories representing the meaning of the source and the target.
    List the images associated from the source $B$ as $B_1, B_2, B_3, ...$ and set them as the initial images of $B$ (denoted as $B_i$).
    Excite the arrows from $B$ to $B_i$ $b_i: B \rightarrow B_i$ and set them as the initial state.
    Similarly, set the initial state (images) for the target $A$.
    Here, the associative relationship ``$A$ is $B$'' occurs, and the arrow $f: A \rightarrow B$ from $A$ to $B$ is elicited.
\item 
    Construct the trivial functor $\mathrm{BMF}: B\backslash \mathcal{C'} \rightarrow A\backslash \mathcal{C'}$, from the coslice category of the source to that of the target.
    Due to the elicitation of the arrow $f$, $f$ and $b_i$ are composed, and the composite arrow $b_i \circ f: A \rightarrow B_i$ from $A$ to $B_i$ occurs.
    In general, $\mathrm{BMF}: b_i \longmapsto b_i \circ f$.
\item 
    Explore a new functor by constructing a natural transformation $\vartheta$ from BMF.
    Consider to where $B_1$ is mapped by the new functor, as an example.
    Acquire the association weights $\mu_{11}, \mu_{12}, \mu_{13}, ...$ from $B_1$ to $A_i$ ($i=1, 2, 3, ...$) from the latent category.
    Each arrow from $B_1$ to $A_i$ is elicited at the probability of $\mu_i$.
    Among the elicited arrows, select the arrow with the largest $\mu_{1i}$.
    If $\mu_{11}$ is the largest,
    the new functor maps the arrow $a_1: A \rightarrow A_1$ to $b_1$,
    and is adopted as an element $\vartheta_{b_1}$ of the natural transformation $\vartheta$ from BMF to $F$.
    If no arrow from $B_1$ to $A_i$ is elicited, then the arrow $b_1$ is mapped to nothing (and $F$ is a partial functor).
\item 
    Carry out this operation for the remaining initial images of the source $B_2, B_3, ...$.
    Suppose $b_2$ correspond to $a_2$ and $b_3$ to $a_3$.
    In this case, $F$ is a functor with $F(b_i) = a_i ~~~ (i = 1, 2, 3)$.
    If there is a natural transformation $\vartheta: \mathrm{BMF} \rightarrow F$, it means that $F$ is  the metaphor is obtained.
    $F$ maps the objects $b_i: B\rightarrow B_i$ in $B\backslash \mathcal{C}'$ (note that they are arrows in $\mathcal{C}$)
    to the objects $a_i: A\rightarrow A_i$ in $A\backslash \mathcal{C}'$.
    The content of the metaphor comprehension is `$B_i$ for $B$' is `$A_i$ for $A$' (for $i=1,2,3$).
\end{enumerate}

\subsubsection{The Relation-based Exploration}

\begin{figure}[tb]
    \centering
    \includegraphics[width=\linewidth]{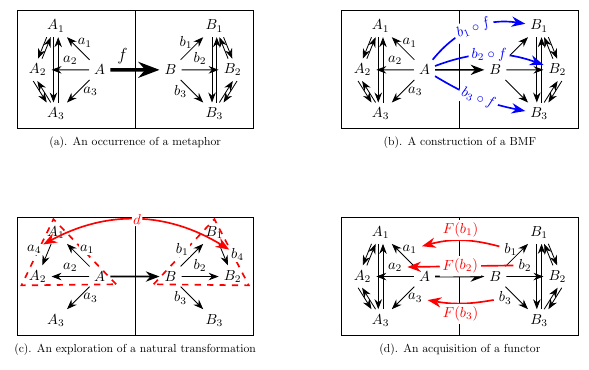}
    \caption{The algorithms of the triangle method}
    \label{fig:triangle}
\end{figure}

\begin{figure}[tb]
    \centering
    \includegraphics[width=.6\linewidth]{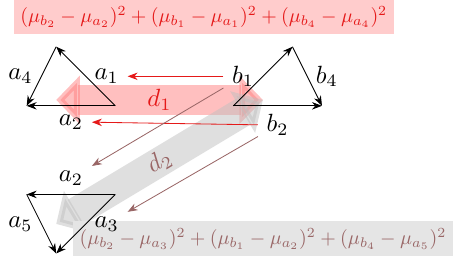}
    \caption{The distance between the triangle structures}
\end{figure}

The relation-based algorithm constructs a mapping from an arrow in the coslice category
($B\backslash \mathcal{C}'$)
to
another ($A\backslash \mathcal{C}'$)\footnote{
The previous relation-based algorithm proposed in \cite{Ikeda-2021-ComputationalImplementationMetaphorComprehension}
output multiple functors, not a single functor.
It made it difficult to compare the algorithm with the object-based one or others.
To address this issue, we developed a more natural relation-based algorithm here.
}.
The arrows in coslice categories are not just arrows in $\mathcal{C}$ which is an association,
but the commutative triangle structures (triplets).
The mapping is searched using the distance between an arrow (triangle) in $B\backslash \mathcal{C}'$ and
arrows (triangles) in $A\backslash \mathcal{C}'$.
It is that the local associative structures of the meaning of the target and source are considered, unlike the object-based method
that ignores any structure.
The algorithm again is described in four steps that correspond to Figure \ref{fig:triangle}-a to \ref{fig:triangle}-d.
 
\begin{enumerate}[\ref{fig:triangle}-a]
\setlength{\itemsep}{0pt}
\setlength{\parskip}{0pt}
\setlength{\parsep}{0pt}
\setlength{\topsep}{0pt}

    \item
    Set the initial state of the coslice categories representing the meaning of the source and the target,
    in the same way as the object-based method (\ref{fig:objective}-a).
    Then, all the arrows between the initial images for each coslice category are elicited.

    \item 
    Construct BMF as in the object-based method (\ref{fig:objective}-b).

    \item
    Explore a functor through a natural transformation from BMF, based on the triangle structures.
    Consider the triangle structure formed by the arrows $b_1, b_2$, and $b_4$ on the source side.
    The associative probabilities in the latent category determine whether the arrows from $B_1, B_2$ to $A_i$ are excited.
    All possible triangle structures that can be constructed from $A_i$ with an arrow from $B_1$, and $A_i$ with an arrow from $B_2$, are the candidates.
    Among them, adopt the triangle structure with the smallest difference in associative probabilities between the corresponding arrows.
    In the example of Figure \ref{fig:triangle}, the difference in the triangle structure is
    $d = \sum_{i = \{1, 2, 4\}}(\mu_{b_i} - \mu_{a_i})^2$.
    If the difference $d$ with the triangle structure formed by $a_1, a_2, a_4$ is the smallest, $b_1$ corresponds to $a_1$ and $b_2$ corresponds to $a_2$.

    \item
    Perform this operation for all the triangle structures on the target side.
    In the example of the Figure \ref{fig:triangle}, there are 
    $3 \times 2 = 6$ structures (as much as the ordered pairs of initial imgaes of the source).
    If we follow the above procedure for all the triangle structures for the source, it is possible that a source side arrow $b_i$ has correspondences with multiple arrows on the target side.
    To determine the correspondence uniquely, compare the difference $d$ of the triangle structure when the correspondence is chosen,
    and adopt the smallest one as an element of the natural transformation from BMF to $F$.
    Finally, we obtain the functor $F$ which is the interpretation of the metaphor.
\end{enumerate}

\subsection{Introducing Stochastic Correspondence}

In this study, we propose a method to replace the deterministic selection of elements of natural transformations in the simulations of the object-based and the relation-based in Ikeda et al, with a probabilistic (softmax) operation.
It is because the selection should reflect the (coslice category) structures that are supposed to represent the meaning of the target and source, while the deterministic selection ignores most of the information of the structures.
In the object-based method, we selected the arrow with largest weight, $\mu$, out of the arrows, excited from the objects of the vehicle's coslice category to that of the target's coslice category in Figure (\ref{fig:objective}-c), as elements of natural transformations based on the associative probabilities.
We replaced this part with a method that stochastically selects elements by a softmax function with each $\mu$ as input, instead of selecting the one in the `greedy' way.
In the relation-based method, we adopted the triangle structure with the smallest difference $d$ as a candidate for natural transformations
in Figure (\ref{fig:triangle}-c), using $d$ in the same softmax selection as the object-based method.

\section{Experiment}

We tested the new TINT algorithms in the same environment as the simulations by \cite{Ikeda-2021-ComputationalImplementationMetaphorComprehension}.
We chose the metaphor `Butterflies are dancers' 
from the metaphor stimulus-interpretation set by \cite{Oka-2019-DevelopmentValidationItemSet}.
We define the association weights in the latent category by the data collected by Ikeda et al. in a cognitive experiment.
In this experiment, participants answered the the associative strength of `$A$ associates $B$'
for all pairs of images on a 5-point scale (1: not associated, ... 5: strongly associated).
After that, the association strength $s$ was converted to the weight $\mu$ by $\mu = 0.05+0.225(s-1)$.
We also used the metaphor interpretation data collected by Ikeda et al. to evaluate the metaphor comprehensions constructed by the TINT algorithms.
The metaphor interpretation data is the data in which people answered the associative strength of how much they agree with the metaphor of the image pair.
We conducted 10,000 simulations of the correspondence from the source initial images to the target's initial images begining with the metaphor.
We compared the simulation results between the previous two algorithms by Ikeda et al. as ``hardmax'' and the proposed method as ``softmax''.

\begin{figure}[tb]
    \centering
    \includegraphics[width=\linewidth]{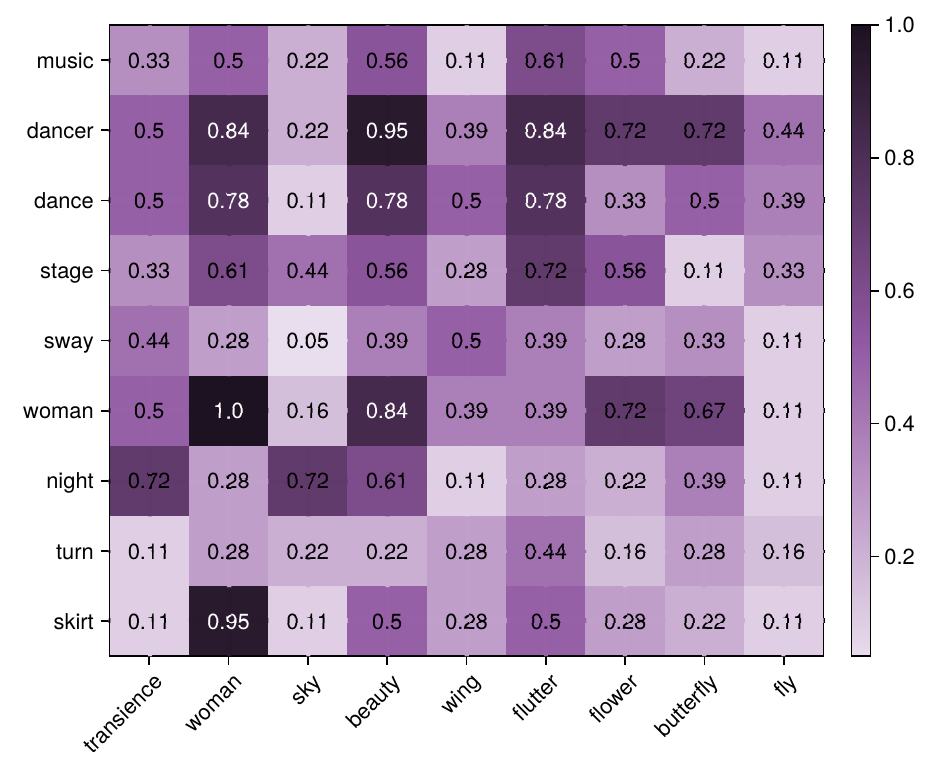}
    \caption{The association weights assigned to the latent category. Only the weights from the initial images of the source to that of the target are shown.}
    \label{fig:assoc_data}
\end{figure}

\section{Result}

We compare the four metaphor comprehension algorithms (relation-based or object-based) $\times$ (deterministic/hardmax or probabilistic/softmax)
in three measures:
data fit,
systematicity of the comprehension,
and
the novelty of the comprehension.
All the results are shown in Figure \ref{fig:result}, with the free parameter $\beta$ (inverse temparature) for the softmax method as a variable on the $x$-axis.

\subsection{Data Fit}
We calculated the rank correlation coefficient between the metaphor interpretation data and the simulation results for each algorithm.
We ranked the initial images of the target by the number of correspondences by TINT for each initial image of the source.
For the metaphor interpretation data, we made a ranking in a similar way, according to the responses by the participants.
We calculated the rank correlation coefficient between them.
The results are shown in Figure \ref{fig:result} as the gray lines.
%
%
The relation-based softmax (gray solid line) shows better datafit than the object-based (gray broken line) or the two methods by Ikeda et al. (gray triangle and circle) for most of the range for $\beta$.

\subsection{Systematicity (the Width of the Functors)}
We considered it is important that the width of the functors as the metaphor comprehension, as the width corresponds to the systematicity in structure-mapping theory~\cite{Gentner-1983-StructuremappingTheoreticalFrameworkAnalogya}.
We defined the width of the functors as the number of objects in the target's coslice category that were mapped to.
In other words, the width is the size of the range of the functor (where, the bigger the size is, the closer the functor is to an injective mapping).
The number of objects in the coslice category is the number of initial images.
The maximum width of the functors is 8. 
The results of comparing the average width for each trial of the simulation are shown
in Figure \ref{fig:result} in orange.
%
%
The width of the functor constructed with the relation-based methods (solid orange curve or orange triangle) exceeded that with the object-based.

\subsection{The Novelty of the Metaphor Comprehension}

\begin{figure}[tb]
    \centering
    \includegraphics[width=\linewidth]{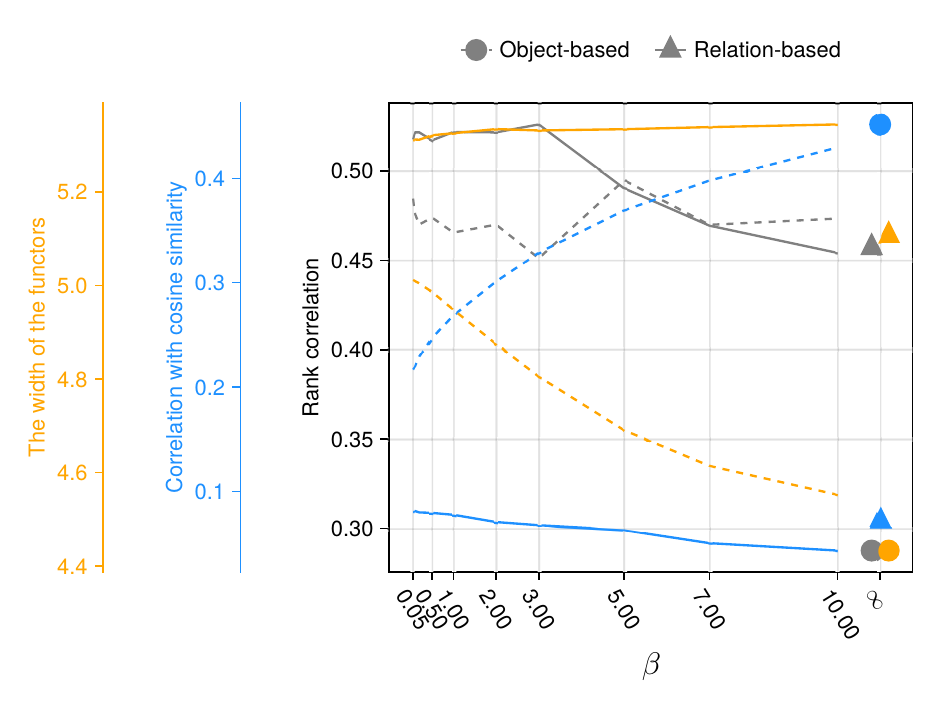}
    \caption{The output of the algorithms evaluated in 1. data fit (gray), 2. systematicity (orange), and 3. novelty (blue).
      The $\beta$ value along the $x$-axis is the inverse temperature parameter for the softmax methods.
      The circles and triangles in three colors are of the algorithms proposed in Ikeda et al. (that adopted the absolute errors, instead of the square errors for the distance between triangle structures, $d$
      ).
      }
    \label{fig:result}
\end{figure}

We examined the novelty of the metaphor comprehension output of TINT algorithms
by comparison with the word embeddings of the images.
If an algorithm can acquire a metaphor comprehension correspondence between pairs of images with lower cosine similarity on the word embeddings, we considered that the comprehension has a relative novelty.
We used word2vec for the word embeddings \cite{Mikolov-2013-DistributedRepresentationsWordsPhrasesa, Mikolov-2013-EfficientEstimationWordRepresentationsa}.
We employed the word embedding model that was pre-trained on the full text of the Japanese Wikipedia \cite{Suzuki-2018-JointNeuralModelFineGraineda} to obtain the distributed representations of the images used in the experiment.
Since the two words `beauty (\textit{utsukushisa})' and `transience (\textit{hakanasa})' were not included in the dictionary of the model, we adopted `beautiful (\textit{utsukushii})'' and `transient (\textit{hakanai})' instead.
Next, we calculated the cosine similarity between the images.
We show the results
in Figure \ref{fig:result} in blue.
%
%
This is a discriminative index that is opposite to the correlation with the metaphor interpretation data.
The relation-based methods showed a much lower correlation coefficient with the word embeddings than the object-based, which means that the relation-based can produce more novel metaphor comprehension.

\section{Discussion}

In Figure \ref{fig:result},
we can see that the proposed softmax method is closer to the metaphor interpretation data than the conventional method hardmax, both in the object-based and the relation-based exploration methods.
In hardmax, most of the correspondences are made with the pairs, initial images of the source and that of the target, that have high association weights.
For example, the initial images of `dancer' and `butterfly' have a common image of `woman' as shown in Figure \ref{fig:assoc_data} (the second column and sixth row).
In this experiment, the associative probability of the identity arrow is set to 1, thus `woman' is always corresponded to `woman' in the object-based hardmax method.
The softmax function in the proposed method could probabilistically disperse the correspondences even in such cases.
We considered that the our method is closer to the metaphor interpretation data due to this feature.
In other words, we suggest that the metaphor comprehension process in humans is based on more probabilistic procedures.
The relation-based method tends to correlate better with humans metaphor interpretation.
In terms of the treatment of analogy and metaphor on structure-mapping theory by Gentner et al., the object-based exploration loosely corresponds to the mapping between attributes, and the relation-based to the mapping between relations, respectively \cite{Gentner-1983-StructuremappingTheoreticalFrameworkAnalogya}.

We compared the broadness of the destination of the objects in the coslice category by the functors.
To express flexible metaphors of humans, the correspondence from initial images of a source to more initial images of a target preferred to that of to particular initial images of a target.
We see in Figure \ref{fig:result} that the relation-based output consistently broader functors compared to the object-based.

Finally, we examined the novelty of the obtained metaphor comprehension using word embeddings.
Figure \ref{fig:result}
shows that the relation-based methods gave far lower correlation with the word embeddings overall.
The results suggested that the relation base could acquire more novel metaphor comprehensions compared to object-based.

\section{Conclusion}

In this paper, we simplified and generalized the  metaphor comprehension algorithms based on TINT.
We proposed two new evaluation indices for comparing the methods, systematicity and novelty of the metaphor comprehension outputs.
The new algorithms proposed in this paper showed better datafit, higher systematicity, and higher novelty.


\bibliographystyle{plain}
\bibliography{ref}

@article{Chalmers-1992-HighlevelPerceptionRepresentationAnalogya,
  title = {High-Level Perception, Representation, and Analogy: {{A}} Critique of Artificial Intelligence Methodology},
  author = {Chalmers, David J. and French, Robert M. and Hofstadter, Douglas R.},
  year = {1992},
  journal = {Journal of Experimental and Theoretical Artificial Intelligence},
  volume = {4},
  number = {3},
  pages = {185--211},
  issn = {13623079},
  doi = {10.1080/09528139208953747}
}

@book{Chater-2019-MindFlata,
  title = {The {{Mind Is Flat}}},
  author = {Chater, Nick},
  year = {2019},
  month = mar,
  publisher = {Penguin},
  urldate = {2025-01-22},
  langid = {american}
}

@article{Falkenhainer-1989-StructuremappingEngineAlgorithmExamples,
  title = {The Structure-Mapping Engine: {{Algorithm}} and Examples},
  shorttitle = {The Structure-Mapping Engine},
  author = {Falkenhainer, Brian and Forbus, Kenneth D. and Gentner, Dedre},
  year = {1989},
  month = nov,
  journal = {Artificial Intelligence},
  volume = {41},
  number = {1},
  pages = {1--63},
  issn = {0004-3702},
  doi = {10.1016/0004-3702(89)90077-5},
  urldate = {2025-02-04}
}

@book{Fong-2019-InvitationAppliedCategoryTheorya,
  title = {An {{Invitation}} to {{Applied Category Theory}}: {{Seven Sketches}} in {{Compositionality}}},
  author = {Fong, Brendan and Spivak, David I.},
  year = {2019},
  month = aug,
  publisher = {Cambridge University Press},
  urldate = {2025-02-02},
  isbn = {978-1-108-71182-1}
}

@article{Fuyama2020a,
  title = {A Category Theoretic Approach to Metaphor Comprehension: {{Theory}} of Indeterminate Natural Transformation},
  author = {Fuyama, Miho and Saigo, Hayato and Takahashi, Tatsuji},
  year = {2020},
  journal = {Biosystems},
  volume = {197},
  number = {March},
  pages = {104213},
  publisher = {Elsevier B.V.},
  issn = {03032647},
  doi = {10.1016/j.biosystems.2020.104213},
  pmid = {32712313}
}

@article{Gentner-1983-StructuremappingTheoreticalFrameworkAnalogya,
  title = {Structure-Mapping: {{A}} Theoretical Framework for Analogy},
  shorttitle = {Structure-Mapping},
  author = {Gentner, Dedre},
  year = {1983},
  month = apr,
  journal = {Cognitive Science},
  volume = {7},
  number = {2},
  pages = {155--170},
  issn = {0364-0213},
  doi = {10.1016/S0364-0213(83)80009-3},
  urldate = {2025-01-22}
}

@article{Holyoak-2018-MetaphorComprehensionCriticalReviewa,
  title = {Metaphor Comprehension: {{A}} Critical Review of Theories and Evidence},
  shorttitle = {Metaphor Comprehension},
  author = {Holyoak, Keith J. and Stamenkovi{\'c}, Du{\v s}an},
  year = {2018},
  journal = {Psychological Bulletin},
  volume = {144},
  number = {6},
  pages = {641--671},
  publisher = {American Psychological Association},
  address = {US},
  issn = {1939-1455},
  doi = {10.1037/bul0000145}
}

@article{Ikeda-2021-ComputationalImplementationMetaphorComprehension,
  title = {Toward Computational Implementation of Metaphor Comprehension Process Based on the Theory of Indeterminate Natural Transformation},
  author = {Ikeda, Shunsuke and Fuyama, Miho and Saigo, Hayato and Takahashi, Tatsuji},
  year = {2021},
  journal = {Cognitive Studies},
  volume = {28},
  number = {1},
  pages = {39--56},
  doi = {10.11225/cs.2020.065}
}

@misc{Mikolov-2013-DistributedRepresentationsWordsPhrasesa,
  title = {Distributed {{Representations}} of {{Words}} and {{Phrases}} and Their {{Compositionality}}},
  author = {Mikolov, Tomas and Sutskever, Ilya and Chen, Kai and Corrado, Greg and Dean, Jeffrey},
  year = {2013},
  month = oct,
  number = {arXiv:1310.4546},
  eprint = {1310.4546},
  primaryclass = {cs, stat},
  publisher = {arXiv},
  doi = {10.48550/arXiv.1310.4546},
  urldate = {2022-10-06},
  archiveprefix = {arXiv}
}

@misc{Mikolov-2013-EfficientEstimationWordRepresentationsa,
  title = {Efficient {{Estimation}} of {{Word Representations}} in {{Vector Space}}},
  author = {Mikolov, Tomas and Chen, Kai and Corrado, Greg and Dean, Jeffrey},
  year = {2013},
  month = sep,
  number = {arXiv:1301.3781},
  eprint = {1301.3781},
  primaryclass = {cs},
  publisher = {arXiv},
  doi = {10.48550/arXiv.1301.3781},
  urldate = {2022-10-13},
  archiveprefix = {arXiv}
}

@article{Oka-2019-DevelopmentValidationItemSet,
  title = {Development and Validation of an Item Set of Simile Interpretations for Metaphor Research},
  author = {Oka, Ryunosuke and Ohshima, Hiroaki and Kusumi, Takashi},
  year = {2019},
  journal = {The Japanese Journal of Psychology},
  volume = {90},
  number = {1},
  pages = {53--62},
  doi = {10.4992/jjpsy.90.17236}
}

@article{Suzuki-2018-JointNeuralModelFineGraineda,
  title = {A {{Joint Neural Model}} for {{Fine-Grained Named Entity Classification}} of {{Wikipedia Articles}}},
  author = {Suzuki, Masatoshi and Matsuda, Koji and Sekine, Satoshi and Okazaki, Naoaki and Inui, Kentaro},
  year = {2018},
  journal = {IEICE Transactions on Information and Systems},
  volume = {E101.D},
  number = {1},
  pages = {73--81},
  issn = {0916-8532, 1745-1361},
  doi = {10.1587/transinf.2017SWP0005},
  urldate = {2025-01-22},
  langid = {english}
}

\end{document}